\documentclass[letterpaper, 10 pt, conference]{ieeeconf}
\IEEEoverridecommandlockouts
\usepackage{algorithmic}
\usepackage{array}
\usepackage{textcomp}
\usepackage{stfloats}
\usepackage{url}
\usepackage{verbatim}
\usepackage{balance}
\usepackage{diagbox}
\usepackage{amsmath,amssymb,amsfonts}
\usepackage{amsmath}
\usepackage{graphicx}
\usepackage{textcomp}
\usepackage{multirow}
\usepackage{booktabs}
\usepackage{soul,color}
\usepackage{CJKutf8}
\usepackage{threeparttable}

\usepackage{subfigure} 
\newlength{\whilewidth}
\newcommand{\tabincell}[2]{\begin{tabular}{@{}#1@{}}#2\end{tabular}}
\usepackage{cite}
\usepackage{algorithm}
\usepackage{graphicx}
\usepackage{etoolbox}

\title{\LARGE \bf
	Learning Goal-oriented Bimanual Dough Rolling Using \\Dynamic Heterogeneous Graph Based on Human Demonstration
} 

\author{Junjia Liu$^{1}$, Chenzui Li$^{1}$, Shixiong Wang$^{1}$,  Zhipeng Dong$^{1}$, \\ Sylvain Calinon$^{2}$, \IEEEmembership{Member, IEEE}, Miao Li$^{3}$, \IEEEmembership{Senior Member, IEEE}, and Fei Chen$^{\dagger 1}$, \IEEEmembership{Senior Member, IEEE}
	\thanks{This work was supported in part by the Research Grants Council of the Hong Kong SAR under Grant 14211723, 14222722, 24209021 and C7100-22GF.}
	\thanks{$^{1}$Junjia Liu, Chenzui Li, Shixiong Wang, Zhipeng Dong,  and Fei Chen are with the Department of Mechanical and Automation Engineering, T-Stone Robotics Institute, The Chinese University of Hong Kong, Hong Kong (e-mail: {\tt\small jjliu@mae.cuhk.edu.hk, czli@mae.cuhk.edu.hk, sxwang@hkclr.hk, zhipengdong@cuhk.edu.hk, f.chen@ieee.org}).}%
	\thanks{$^{2}$Sylvain Calinon is with the Idiap Research Institute, Martigny, Switzerland  (e-mail: {\tt\small sylvain.calinon@idiap.ch}).}
		\thanks{$^{3}$Miao Li is with the School of Microelectronics and the
			Institute of Technological Sciences, Wuhan University, Wuhan, China  (e-mail: {\tt\small miao.li@whu.edu.cn}).}
	\thanks{$^\dagger$Corresponding authors}
}

\begin{document}
	\maketitle
	
	\begin{abstract}
Soft object manipulation poses significant challenges for robots, requiring effective techniques for state representation and manipulation policy learning. State representation involves capturing the dynamic changes in the environment, while manipulation policy learning focuses on establishing the relationship between robot actions and state transformations to achieve specific goals. To address these challenges, this research paper introduces a novel approach: a dynamic heterogeneous graph-based model for learning goal-oriented soft object manipulation policies. The proposed model utilizes graphs as a unified representation for both states and policy learning. By leveraging the dynamic graph, we can extract crucial information regarding object dynamics and manipulation policies. Furthermore, the model facilitates the integration of demonstrations, enabling guided policy learning. To evaluate the efficacy of our approach, we designed a dough rolling task and conducted experiments using both a differentiable simulator and a real-world humanoid robot. Additionally, several ablation studies were performed to analyze the effect of our method, demonstrating its superiority in achieving human-like behavior.
	\end{abstract}
	
\section{Introduction}
In the context of domestic scenarios, the manipulation of soft objects using dual arms is a common requirement. Numerous studies in the field of robotics have focused on bimanual activities to advance the application of robots in service-oriented tasks. Examples include folding clothes \cite{yin2021modeling}, coffee stirring \cite{sun2022mixline}, and cooking with stir-fry \cite{liu2022robot}. These tasks necessitate accurate state estimation of soft objects and the execution of coordinated bimanual movements. It is crucial to employ a reliable state representation method that can robustly capture the changes in the state of soft objects despite disturbances from environmental factors and sensor limitations. Additionally, an important practical consideration is the ability to transfer policies learned in simulation to real-world scenarios, bridging the significant gap between simulated and real environments. Such capability would greatly facilitate the practical implementation of simulator-based learning methods in real-world applications.
\begin{figure}[t]
	\centerline{\includegraphics[width=0.75\linewidth]{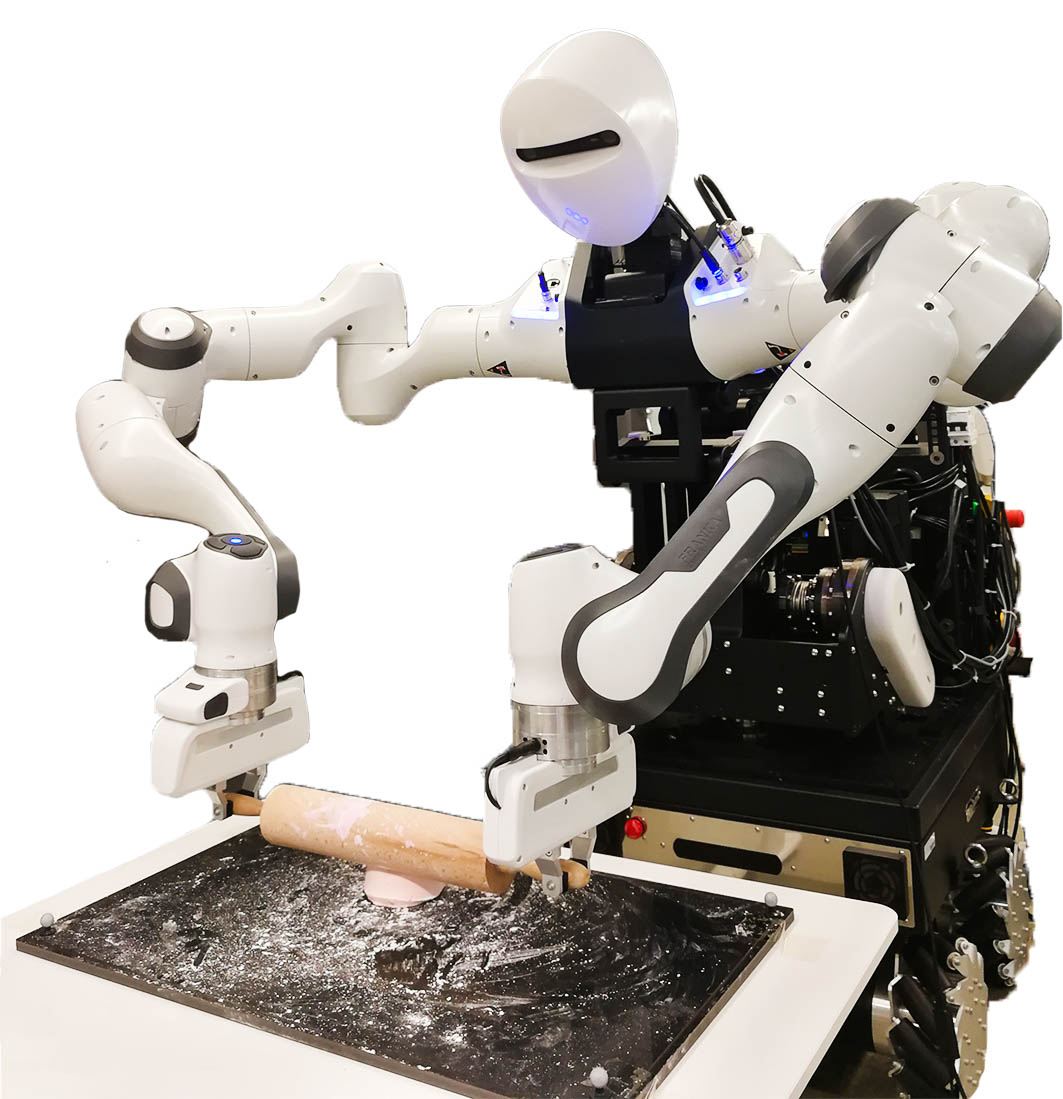}} 
	\caption{Perform a bimanual dough rolling task with a rolling pin on a CURI robot by using the pre-trained dynamic heterogeneous graph model.}
	\label{dough_task}
		\vspace{-1.em}
\end{figure}
After obtaining a compact and concise state information, robots are required to learn the bimanual manipulation policy to achieve specific goals. However, the complexity of movements and the requirement for scene generalization make simple logic programming inadequate for such tasks. In recent decades, two mainstream solutions have emerged: Reinforcement Learning (RL) and Learning from Demonstration (LfD). However, both approaches have their limitations. Demonstrations provide direct examples of expert policies, enabling the robot to mimic behavior in a supervised manner. This approach restricts the robot's ability to learn object state representation and discover new skills due to the limited amount of available data. On the other hand, RL-based methods allow for extensive interaction with the environment but often struggle to learn complex or long-horizon tasks from scratch. In summary, state representation and policy learning are the primary factors that restrict robots' ability to acquire human-like skills, particularly in the domain of soft object manipulation. Therefore, this paper aims to address these two challenges simultaneously and in a natural manner by proposing a suitable approach.

\begin{figure*}[ht]
	\centerline{\includegraphics[width=1\linewidth]{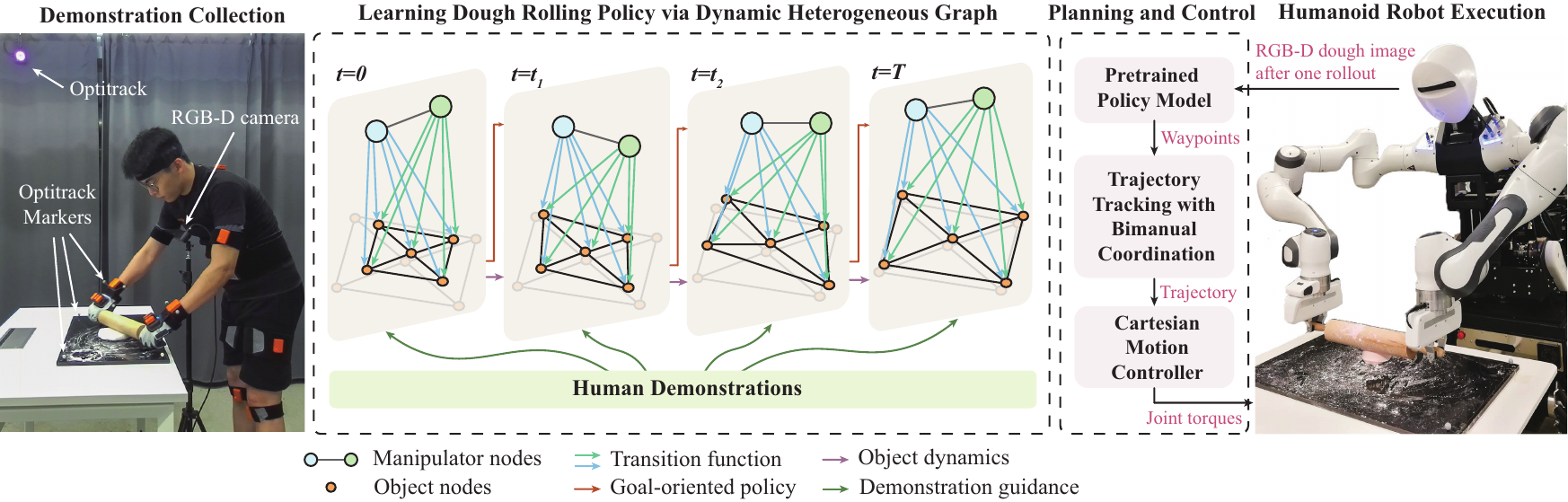}} 
	\caption{The proposed pipeline of soft object manipulation, which contains two parts: dynamic heterogeneous graph policy learning in the simulator with the guidance of demonstration, and planning and control for deploying on the real-world humanoid robot.}
	\label{pipeline}
		\vspace{-1.em}
\end{figure*}

Soft object state estimation and representation, a crucial challenge in the field of robotics, has garnered significant attention in the realm of Computer Graphics. Huang et al. introduced PlasticineLab, a state-of-the-art differentiable physics benchmark, supported by the DiffTaichi system \cite{hu2019difftaichi}. PlasticineLab utilizes the MLS-MPM algorithm to simulate plasticine-like objects. Notably, the differentiable physics framework provides analytical gradient information, enabling supervised learning with gradient-based optimization. The versatility of PlasticineLab has led to numerous studies focusing on soft object manipulation. For instance, Shi et al. proposed the utilization of graph neural networks (GNNs) to learn the particle-based dynamics of soft objects for model-based planning \cite{RoboCraft}. These advancements highlight the potential of employing differentiable physics simulations and graph neural networks to enhance soft object state estimation and manipulation strategies in the field of robotics.

Soft object manipulation policy research primarily revolves around two approaches: direct imitation from human demonstration and policy transfer from simulators. The former involves supervised learning techniques, such as Dynamic Movement Primitives (DMP) \cite{schaal2006dynamic}, TP-GMM \cite{calinon2016tutorial}, and deep learning models \cite{ho2016generative}, to obtain optimal policies by fitting data from demonstrations. The latter leverages reinforcement learning (RL) to train robots in simulators like Habitat \cite{savva2019habitat}, Mujoco \cite{todorov2012mujoco}, pyBullet, DiffTaichi \cite{hu2019difftaichi}, and PlasticineLab \cite{huang2021plasticinelab}. While demonstration provides expert policies and mimics human behavior \cite{liu2022robot}, it may have limitations in learning object state representation and discovering new skills due to limited data. Simulators bridge the gap with reality but struggle with policy transfer. Combining demonstration and simulation allows for the integration of their strengths and mitigates their weaknesses \cite{liu2020efficient}.

Relational learning offers a unified solution to the above challenges of soft object manipulation. Geometric information of soft objects can be represented as a graph, where robot end-effectors are abstracted as manipulator nodes to capture the relationship between manipulation and state change. By modeling continuous manipulation behavior as the evolution of dynamic graphs, we can learn robot behavior oriented towards object state changes. Demonstration data can also be represented as dynamic graphs, enabling seamless integration with simulator-based policy learning and overcoming differences between reality and simulation. This approach provides a comprehensive framework for robust and adaptable soft object manipulation in dough rolling task.

This paper presents several key contributions. Firstly, we propose a \textbf{D}ynamic heterogeneous \textbf{G}raph based model for learning \textbf{G}oal-oriented de\textbf{form}able manipulation policy (\textbf{DGform}). This model adopts a unified graph representation for both state information and policy learning, incorporating guidance from human demonstrations. Secondly, we provide a comprehensive pipeline for deploying the pre-trained policy on a real-world humanoid robot, specifically focusing on the task of dough rolling. Lastly, we conduct ablation studies and generalization analysis to demonstrate the superior performance of our proposed method in learning human-like bimanual skills. 

\section{Dough Rolling Task}
%
%

To evaluate our method, we tackle the challenging task of bimanual dough rolling. Previous  dough rolling research includes Figueroa et al., who employed a hierarchical framework for automatic task segmentation and represented human demonstrations using action primitives \cite{figueroa2016learning}. However, such movement primitive-based methods have limitations in understanding the relationship between robot motions and object state changes. Calinon et al. discussed generalization in learning from few demonstrations, presenting a task-parameterized mixture model and comparing it with other approaches \cite{calinon2013improving}. It primarily focused on determining the rolling direction rather than accomplishing the complete dough rolling task.

In recent dough rolling research, there has been a growing reliance on differential simulators. Diffskill, for instance, introduced a method for learning skill abstraction in sequential dough manipulation tasks. It employed a differentiable simulator to generate simulated demonstrations and trained a supervised policy for each dough manipulation primitive skill \cite{lin2022diffskill}. Another notable work proposed a Generative Pre-trained Transformer (GPT) \cite{radford2019language} based world model (SoftGPT) for learning the dynamics of dough from a differential simulator \cite{liu2023softgpt}. This model facilitated efficient learning of diverse dough manipulation tasks including dough rolling, but it only conducted experiments in the simulator.




\section{Soft Object Manipulation Policy Learning via Dynamic Heterogeneous Graph}
In this section, we will introduce the key concepts and components used in our proposed methods. The definitions of important symbols are provided in Table \ref{term}. The overall soft object manipulation pipeline is illustrated in Fig. \ref{pipeline}, consisting of two main parts: (1) dynamic heterogeneous graph policy learning in the simulator with the guidance of demonstration, and (2) planning and control for deploying the pre-trained model on the real-world humanoid  robot. 



\subsection{Represent soft objects and manipulator as a heterogeneous graph}\label{IIIA}
Recent studies have started utilizing graph-structured data and graph neural networks to model soft object manipulation policy learning. For instance, some research has focused on predicting action effects of a bag using graph-based approaches \cite{weng2021graph}, learning the dynamics of soft objects \cite{RoboCraft}. 

Graphs serve as efficient and compact data structures for representing non-Euclidean data, capturing both semantic and geometric information through their nodes and edges. To address the challenge of abstracting the state of a dough object, which lacks distinct characteristic points unlike soft objects such as clothes \cite{yin2021modeling}, we employ a robust solution that minimizes interference with subsequent policy learning. In this study, we utilize standard image processing techniques to achieve this abstraction. Initially, we segment the dough from the operation board using a color threshold and subsequently determine the center of the segmentation contour. From this contour center, we establish four lines with equal angular spacing and calculate eight focal points between them and the contour. These focal points serve as position features for the object's boundary nodes. Alongside the center node, we construct a 9-node $\mathcal{V}_o$ graph to represent the object's state. Each node in the graph is associated with a position feature and a corresponding depth feature obtained from depth observations. To provide clarity, Fig. \ref{pipeline} illustrates an example with five object nodes. The graph includes edges connecting all object boundary nodes to the center node, while the boundary nodes are connected solely to their two immediate neighbors. This graph-based representation facilitates effective modeling of the dough object's state and its subsequent utilization in the research field of robotics.

\begin{table}[ht]
	\centering
	\caption{Definition of Symbols} 
	\label{term}
	\begin{threeparttable}
		\begin{tabular}{ll}
			\toprule 
			Symbols  &  Definition \\
			\midrule
			$D$ & The dimension of state \\
			$O$ & The order of the controller \\
			$T$ & Time horizon, $t\in [0, T]$ \\
			$o_t$ & The state of the object at time $t$ \\
			$o_g$ & The goal state of the object \\
			$a_t$ & The Cartesian pose of robot dual end-effectors \\
			$\mathcal{V}_m, \mathcal{V}_o$ & Manipulator nodes and object nodes \\
			$\mathcal{E}_{mo}, \mathcal{E}_{oo}$ & Edges between $V_m$ and  $V_o$, and among $V_o$s \\
			${X_\mathcal{V}}_t, {X_\mathcal{E}}_t$  & \tabincell{l}{The attribute matrices of nodes and edges\\ at time $t$} \\
			$\mathcal{G}_t=(\mathcal{V}_t, \mathcal{E}_t)$  & The static graph at time $t$ \\
			\tabincell{l}{$X\mathcal{G}_t=(\mathcal{V}_t, \mathcal{E}_t,$\\  $\quad\quad {X_\mathcal{V}}_t, {X_\mathcal{E}}_t)$}  & The static graph at time $t$ with attributes \\
			$S\mathcal{G}_t, SX\mathcal{G}_t$  & The static object sub-graph at time $t$ \\
			$\boldsymbol{\zeta}={\lbrack\zeta_1^\intercal, \dots, \zeta_T^\intercal\rbrack}^\intercal $ & \tabincell{l}{Demonstrated bimanual trajectory, \\each $\zeta_t={[{\zeta_t^{(l)}}^\intercal, {\zeta_t^{(r)}}^\intercal]}^\intercal\in \mathbb{R}^{2DO}$}  \\
			$SX\mathcal{G}_{\zeta_t}$ & Demonstrated static object sub-graph at time $t$    \\
			\bottomrule[1pt]
		\end{tabular}
	\end{threeparttable}
		\vspace{-1.em}
\end{table}

As shown in the graph block in Fig. \ref{pipeline}, except for the object sub-graph $S\mathcal{G}_t$, the dual end-effectors are added as \textit{manipulator nodes}, together constructing the complete heterogeneous graph $\mathcal{G}_t$. Heterogeneous graph refers to graphs that contain different types of nodes and also different types of edges between nodes. Therefore, a heterogeneous graph representation allows us to learn more relationships as we expected. The manipulator node features are their poses with respect to the board coordinate system. The edges between the manipulator and object nodes are fully connected.

\subsection{Learn state change as the evolution of the dynamic graph}\label{IIIB}
After gaining the graph abstraction of object and manipulator states at each time step, we then consider describing the dynamic evolution of the state and the corresponding manipulation trajectory. Each static graph $\mathcal{G}_t$ is a snapshot of the dynamic graph evolving from time $0\sim T$. Thanks to the nature of the heterogeneous graph, we can define several different types of graph convolution operations on this graph. As shown in Fig. \ref{pipeline}, we learn the goal-oriented policy (Equ. \ref{Za}), transition function and object dynamics (Equ. \ref{Zb}) and value function (Equ. \ref{Zc}) in a unified dynamic heterogeneous graph model.
\begin{subequations}
	\begin{align}
		\hat{a}_{t+1} &\leftarrow \hat{X}_{\mathcal{V}_{m, t+1}} = \pi_{\text{GraphConv}}(X_{\mathcal{V}_{m, t}}, X_{\mathcal{V}_{o, t}}|  X_{\mathcal{SG}_{g}}) \label{Za}\\
		\hat{S\mathcal{G}}_{t+1} &\leftarrow \hat{X}_{\mathcal{V}_{o, t+1}} = Tr_{\text{GraphConv}}(X_{\mathcal{V}_{m, t}}, X_{\mathcal{V}_{o, t}}) \label{Zb} \\
		v&\leftarrow \mathrm{V}(\hat{X}_{\mathcal{V}_{m, t+1}}, \hat{X}_{\mathcal{V}_{o, t+1}}) \label{Zc}
	\end{align}
\end{subequations}

The goal-oriented policy takes the graph abstraction of the goal state (the semitransparent graphs in Fig. \ref{pipeline}) as a condition and learns to obtain the hidden feature of the manipulator node at the next time step through the graph feature of the current state. Similarly, another Graph Convolutional Network (GCN) \cite{kipf2016semi} is set for predicting the next hidden feature of object nodes. These hidden features are transformed to the next action $\hat{a}_{t+1}$ and next object state prediction $\hat{S\mathcal{G}}_{t+1}$ by two additional feed-forward networks. The value of the current policy is estimated upon these hidden features as well. Specifically, we first perform a mean aggregation of these heterogeneous hidden features and then feed it to another feed-forward network.

Since several sub-tasks exist simultaneously, the loss function is composed of several parts. Referring to RL methods, we calculate the advantage function based on the value estimation and the actual return and generate the value loss and the action loss. Another term is a supervised loss for learning an accurate dynamic model.
\begin{equation}
	\begin{aligned}
		\mathcal{L}_t(\theta) = \mathbb{E}[\mathcal{L}^{CLIP}_t+c_1\mathcal{L}^{VF}_t+c_2\mathcal{L}^{Dyn}_t-c_3S[\pi](\hat{S\mathcal{G}}_{t})]
	\end{aligned}
\end{equation}

%
%

\subsection{Demonstration guidance during the policy learning}\label{IIID}
The goal-oriented policy has shown the superiority of integrating the goal state in a compact dimension via graph representation. Another advantage is combining the fully explore-based methods with learning from demonstration methods, adding the demonstration as a policy learning guidance. This topic is studied in the offline RL field.  
The intention of CQL is to lower-bound the actual value of Q function by introducing a conservative term, avoiding the common value overestimation problem in offline RL. This term minimizes values under an appropriately chosen distribution over state-action tuples and then further tightens this bound by incorporating a maximization term over the data distribution \cite{kumar2020conservative}. Since DGform has not been extended to an actor-critic form, a state-action value estimation is not available. Thus, we adopt the current policy distribution to evaluate demonstration state-action pairs.
\begin{equation}\label{no_lag} 
	\begin{split}
		\begin{aligned}
			\mathcal{L}_t^{Imi}(\boldsymbol{\zeta}|\theta) = - \sum_{\boldsymbol{\zeta}} \pi_\theta(\zeta_t|SX\mathcal{G}_{\zeta_t})
		\end{aligned}
	\end{split}
\end{equation}

As we expect, demonstration guidance should not be a strong supervision constraint but should leave room for exploration. Therefore, we further add a self-adjusting term to convert Equ. \ref{no_lag} to a Lagrange version.
\begin{equation}\label{lag} 
	\begin{split}
		\begin{aligned}
			\hat{\theta}^{k+1} \leftarrow &\arg \min _{\theta}\max_{\alpha} \mathbb{E}\Big[\left(\alpha \mathcal{L}_t^{Imi}(\boldsymbol{\zeta}|\theta) -\tau\right) + \mathcal{L}^{CLIP}_t\\ &+c_1\mathcal{L}^{VF}_t+c_2\mathcal{L}^{Dyn}_t-c_3S[\pi](\hat{S\mathcal{G}}_{t})\Big]
		\end{aligned}
	\end{split}
\end{equation}
where $\tau$ is the imitation threshold, $\alpha$ is a self-adjusting Lagrange coefficient of conservative term. More specifically, if the difference between the expected value and the threshold $\tau$ is less than 0, then $\alpha$ will tend to 0. Conversely, $\alpha$ will increase, and the weight of the conservative term in the entire loss calculation will also increase. This version intends to adaptively change the supervision intensity according to the deviation between the rollout and demonstration actions while leaving room for exploration in the vicinity of the demonstration.

\subsection{Model-based rollout by the learned object dynamics}\label{IIIC}
Although directly using image processing methods like segmentation to represent soft objects in a 2-dim graph simplifies our reliance on the visual front-end, it also brings some practical problems. It is challenging to represent the state of the dough by this method when occluding by other objects, like the rolling pin and robot hands. We certainly can also use methods similar to RoboCraft \cite{RoboCraft} to analyze the occlusion of objects by setting up cameras with multiple views in their experiments. However, information incompleteness is the most common in real scenarios. A more appropriate and robust solution is to predict actions over a period of time through the learned model when the state is unknown for part of the period. Formally, for a short horizon $t\in t_k\sim t_k+H$, since the step-by-step interaction with the environment is not available, the proposed model alternately generates the features of object and manipulator nodes.
\begin{subequations}
	\begin{align}
		\hat{a}_{t+1} &\leftarrow \hat{X}_{\mathcal{V}_{m, t+1}} = \pi(\hat{X}_{\mathcal{V}_{m, t}}, \hat{X}_{\mathcal{V}_{o, t}}|  X_{\mathcal{SG}_{g}})  \\
		\hat{S\mathcal{G}}_{t+1} &\leftarrow \hat{X}_{\mathcal{V}_{o, t+1}} = Tr(\hat{X}_{\mathcal{V}_{m, t}}, \hat{X}_{\mathcal{V}_{o, t}}) 
	\end{align}
\end{subequations}

Then we can get a sequence of predicted actions $\hat{\mathbf{a}}=[\hat{a}_{t_k} \dots \hat{a}_{t_k+H}]$ and their corresponding effect prediction. The simulator performs these actions sequentially to obtain the actual feedback (return and next state sequence). Furthermore, the loss function needs to be modified to a period form accordingly.

\subsection{Bimanual Trajectory Planning}
The action output of DGform policy is path waypoints of dual end-effectors, which is inadequate to perform directly on robots with high frequencies (1000Hz in control loop for the humanoid robot in this paper). Thus, a trajectory planning module is required. Besides, since the dual end-effectors has a fixed relative relationship, coordination needs to be considered in the trajectory planning.  Here we use Bimanual Relative Parameterization (Bi-RP) \cite{liu2023birp} for modeling the coordination and generating bimanual trajectories with Linear Quadratic Tracking (LQT) \cite{calinon2017learning}. 

Formally,  the coordination is defined as another frame that takes the trajectory of the other arm as dynamic task parameters and represents the relative relationship as GMM. The relative motion between dual end-effectors is described as $\boldsymbol{\zeta}^{(c)}=\boldsymbol{A}_{c, t}^{-1}\left((\boldsymbol{\zeta}^{(l)}-\boldsymbol{\zeta}^{(r)})-\boldsymbol{b}_{c, t}\right)$ and represented by $\{\pi^{(k)},\boldsymbol{\mu}_c^{(k)},  \boldsymbol{\Sigma}_c^{(k)}\}_{k=1}^K$. For each end-effector, the task-specific GMM obtained by PoE 
\begin{equation}
	\begin{small}
		\begin{aligned}
			\mathcal{N}\left(\boldsymbol{\hat{\nu}}^{(k)}, \boldsymbol{\hat{\Gamma}}^{(k)}\right) \propto \prod_{j=1}^P \mathcal{N}\left(\boldsymbol{\nu}_{j}^{(k)}, \boldsymbol{\Gamma}_{j}^{(k)}\right) \cdot \mathcal{N}\left(\boldsymbol{\nu}_{c}^{(k)}, \boldsymbol{\Gamma}_{c}^{(k)}\right)
		\end{aligned}
	\end{small}
\end{equation}
where $\boldsymbol{\nu}_{c}^{(k)}=\boldsymbol{A}_{c, t}\boldsymbol{\mu}_{c}^{(k)}+\boldsymbol{b}_{c, t}, \boldsymbol{\Gamma}_{j}^{(k)}=\boldsymbol{A}_{c, t}\boldsymbol{\Sigma}_{j}^{(k)}\boldsymbol{A}_{c, t}^\top$.  $\boldsymbol{A}_{c, t}, \boldsymbol{b}_{c, t}$ are transformation matrices. $P$ is the number of reference frames.  More details can be found in \cite{liu2023birp}.

\begin{figure*}[ht]
	\centerline{\includegraphics[width=1\linewidth]{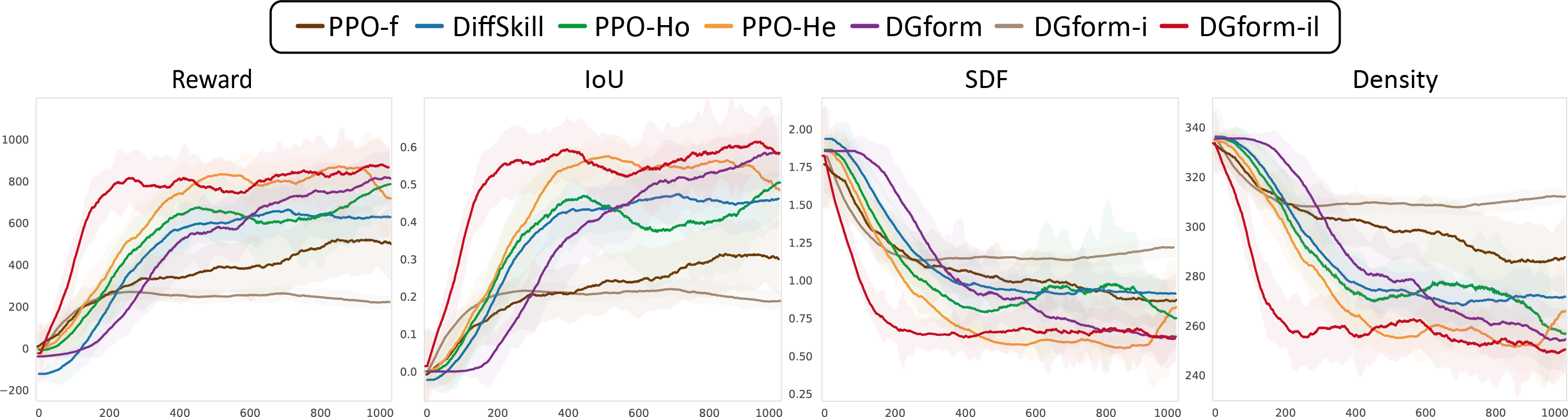}} 
	\caption{The learning performance between baselines and DGform is compared from rewards to three evaluation metrics (SDF, density, and IoU) with the help of the differentiable simulator. The comparison between PPO-based methods demonstrate the effective of graph representation, while three DGform variants show the feasibility of learning policy via graph and the role of demonstration guidance and Lagrangian terms.} 
	\label{learning}
\end{figure*}

\begin{figure*}[ht]
	\centerline{\includegraphics[width=1\linewidth]{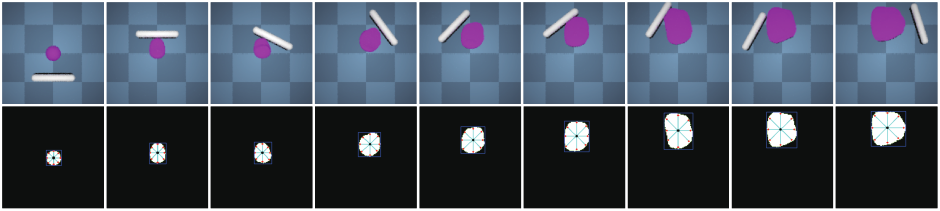}} 
	\caption{An example of DGform inference in the simulator. The short horizon $H$ in the simulation is set to be $50$. The rolling pin will be reset to its initial state after each short horizon and inferred based on the current deformation and the goal state. The first row shows the sequential rolling pin movements sample, and the second is the corresponding visualization of object graph abstractions. Red nodes are boundary nodes, and black nodes are center nodes.}
	\label{sim}
		\vspace{-1.em}
\end{figure*}

\section{Experiments}

\subsection{Setup}
\textbf{Simulation} The simulation environment is built on top of PlasticineLab \cite{huang2021plasticinelab}, with a particle-based dough and a rolling pin. We set up cameras with multiple views to show how the rolling pin and dough change in 3-dim space, while the RGB-D observation for policy learning is obtained from the top-down view. To shorten the learning process while maintaining graph abstraction performance, the resolution of the camera was set to (128, 128).

\textbf{Real-world experiments} The pre-trained model was deployed on the Collaborative dUal-arm Robot manIpulator (CURI) robot, a self-designed bimanual human-like robot equipped with two 7 DoF arms and a 3 DoF torso. CURI utilizes a Cartesian motion controller to execute desired Cartesian space end-effector trajectories by generating joint torques. In real experiments, feedback is provided through first-person perspective observations captured by a single ZED 2i camera embedded in CURI's head. 

\textbf{Evaluation metric} Since the full state of particles is available in the simulator, we compare the Signed Distance Field (SDF), density, and Intersection over Union (IoU) between the current state and the target state.

\textbf{Network details} The model consists of two-layer heterogeneous Graph Convolutional Neural network and a two-layer MLP-based policy learning network combining CQL with PPO, as described in \ref{IIID}.

The preprocessing of the multimodal demonstration data and the implementation of the LQT method depend on the Rofunc package, a full-process python package for robot learning from demonstration published by our lab \cite{liu2023rofunc}. It provides numerous interfaces for demonstration collection and processing, baseline LfD methods, planning methods, and Isaac Gym-based simulators.

\subsection{Demonstration collection}
To incorporate human demonstration during policy learning, we adopt a comprehensive multimodal data collection approach, including visual and motion capture information. Since our focus is on generating bimanual Cartesian trajectories for the rolling pin handles, the whole-body human motion is simplified to dual hand motions. The spatial relationship between the demonstrator hands and the worktop was recorded using the Optitrack system. Specifically, two groups of markers are attached to the demonstrator's wrists, while four markers are fixed on the board to indicate the position of worktop, as shown in the left side of Fig. \ref{pipeline}. To capture the dynamic changes in the dough state, we employ ZED 2i cameras, which provide RGB-D data. By synchronizing and abstracting the data from these sensors as a sequence of heterogeneous graphs, following the techniques described in Section \ref{IIIA}, we create a demonstration dataset. This dataset is utilized in batch to calculate the imitation loss $\mathcal{L}_t^{Imi}$ as defined in Equation \ref{no_lag}. During the demonstrations, the demonstrator was instructed to roll the dough towards a large circular shape within 10 rollouts.


\begin{figure*}[ht]
	\centerline{\includegraphics[width=1\linewidth]{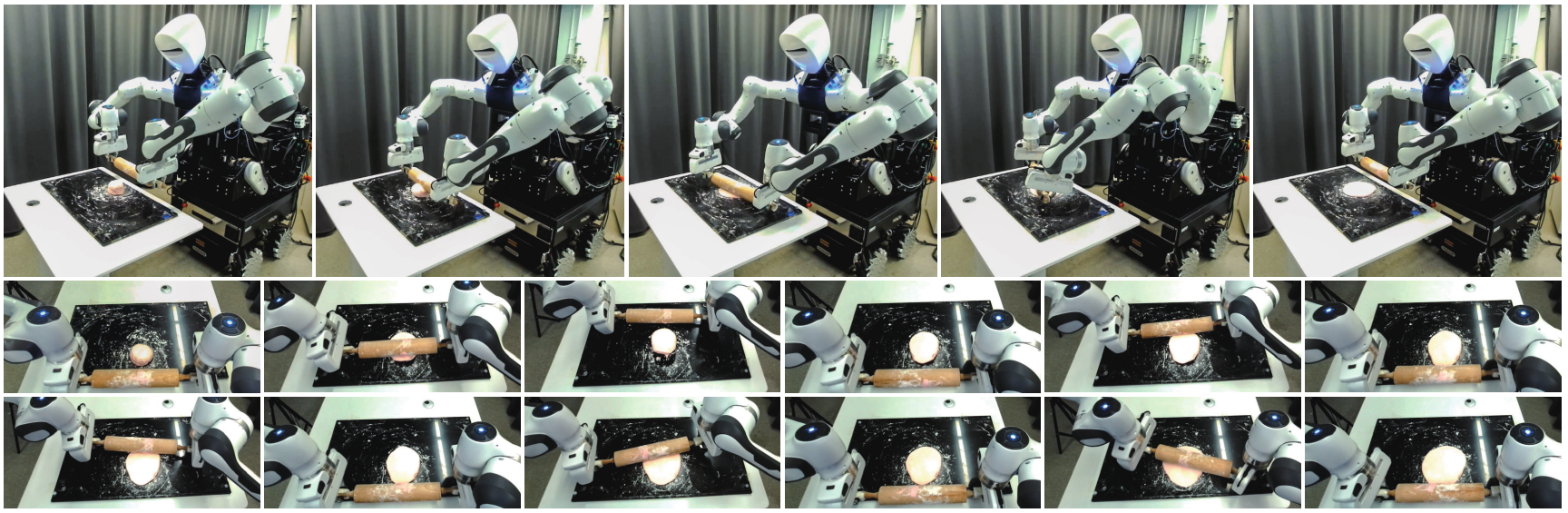}} 
	\caption{The real robot experiment consists of five rollouts. Each time at the initial pose, the camera in CURI's head captured an RGB-D image of the dough, then it was converted into a heterogeneous graph through graph abstraction and passed as input to the pre-trained DGform model. The model generated a path with 50 waypoints, which was then transformed into a smooth dual-arm trajectory with 10,000 points by bimanual LQT. The trajectories were executed by the Cartesian motion controller embedded in the CURI. The first row shows the recording from a third perspective, while the last two are observations from the first-person perspective camera.}
	\label{real}
\end{figure*}

\begin{table}
	\centering
	\caption{Average Performance of Different Scenes} 
	\label{table2}
	\begin{threeparttable}
		\begin{tabular}{p{1.3cm}p{1.3cm}p{1.3cm}p{1.3cm}p{1.3cm}}
			\toprule[1pt]
			Methods 			& Reward 	& IoU & SDF  & Density   \\
			\midrule
			PPO-f$^*$ 			& - 		& - 	 & - 	  & -  \\
			DiffSkill 				& 63.7		& 0.0733 & 1.5307 & 327.9\\
			PPO-Ho 				& 509.3 	& 0.3883 & 0.9056 & 284.5 \\
			PPO-He 				& 629.5 	& 0.4203 & 0.6422 & 271.9 \\
			\textbf{DGform}$^\dagger$	& \textbf{756.1} 	& \textbf{0.5573} & \textbf{0.6223} & \textbf{266.4}\\
			\bottomrule[1pt]
		\end{tabular}
		\begin{tablenotes}
			\footnotesize
			\item[$ * $] In any case, the full particle states are the same.
			\item[$ \dagger $] The three variants of DGform share the same state representation, so here  only the vanilla DGform is added as a representative.
		\end{tablenotes}
	\end{threeparttable}
	\vspace{-2.5em}
\end{table}

\subsection{Policy learning performance}
An inference example of the DGform rolling policy is shown in Fig. \ref{sim}, the first row shows the sample of sequential rolling pin movements, and the second row is the corresponding visualization of graph abstractions.

\textbf{Baselines} The learning performance comparison is mainly divided into two categories, exploring the impact of state representation (full particle state, RGB-D and graph) and the impact of policy learning methods (RL, Lfd, or mixed via DGform).

\begin{itemize}
	\item \textit{PPO with the full state (PPO-f)}: Use full particle states as the observation of PPO.
	\item \textit{PPO with RGB-D state (DiffSkill)}: Use RGB-D images from a top-down view as the observation of PPO, like the primitive skill learning model used in DiffSkill \cite{lin2022diffskill}.
	\item \textit{PPO with Homogeneous Graph state (PPO-Ho)}: Convert the RGB-D image to a homogeneous object sub-graph, and use it as observations of PPO.
	\item \textit{PPO with Heterogeneous Graph state (PPO-He)}: Convert the RGB-D image to a heterogeneous graph, and use it as observations of PPO.
\end{itemize}

Except for the PPO-based baselines, we provide three variants of DGform: vanilla DGform, DGform with imitation (\textit{DGform-i}), and DGform with imitation and Lagrange (\textit{DGform-il}). It is worth mentioning that all baselines do not have a model-based module and need to interact with the environment step-by-step. In contrast, DGform and its variants can rollout a trajectory in a period with the help of the learned model (Sec. \ref{IIIC}). Without immediate interaction, the performance might be influenced, but it will also be robust and practical in real-world scenarios.

From Fig. \ref{learning}, we can figure that the performance is better when reducing the state dimension from full particle state or RGB-D to compact graph state. Among them, PPO with full particle state obtains the worst performance because it is intractable for the network to process tens of thousands of dimensional inputs. In the graph-based method, the heterogeneous graph adds the information of the manipulator. It thus has additional graph convolutional networks to acquire more relations implied in the soft object manipulation. Comparing these four PPO-based methods demonstrates that graph abstraction is sufficient to represent the deformation of objects and is an effective way of state representation. 

On the other hand, policy learning with dynamic graph (DGform) surpasses the PPO-based methods at a slow but robust pace. The inefficiency of this method is primarily affected by the non-interactive model-based rollout since there is little actual interaction with the environment. This makes sense for deploying real robots but also runs the risk of difficult convergence. Thus, we introduce the demonstration guidance, which shows a solid guiding effect in the early stage of learning, driving the policy in the direction of demonstration and avoiding wasting time in meaningless state-action pairs. However, DGform-i without Lagrange cannot converge to normal performance. This is likely constrained by demonstration data taken from different environments. This issue has been addressed by modifying it to a Lagrange version, evaluating the weight of demonstration guidance items in policy updates according to the actual policy dynamically. Finally, the Lagrange version DGform with demonstration guidance achieves the best performance most efficiently.

\subsection{Generalization analysis}
Another essential indicator for evaluating robot skill learning is whether it is easy to transfer to unseen scenarios. With the help of differentiable simulators, we designed generalization experiments with different initial states or camera positions. These two variables are generated randomly in a reasonable range. Their average values of evaluation metrics are shown in Table \ref{table2}. The results were inferred from pre-trained baselines and DGform models. It is easy to figure out that methods with graph state representation have better transfer ability. Moreover, the manipulation performance of DGform does not change much as the environment changes.

\subsection{Real robot dough rolling}
The process of deploying a pre-trained simulated policy on a real-world robot is depicted in the right side of Fig. \ref{pipeline}. At the start of the experiment, the initial state of the dough is captured by the embedded ZED 2i camera in CURI's head, which provides an RGB-D image. This information is abstracted as an object subgraph using the technique described in Section \ref{IIIA}. To determine the relative pose of the robot end-effectors with respect to the center of the worktop, we employ coordinate transformations between the robot hands and the camera. To ensure consistency with the simulator, the poses of the robot end-effectors are set to be the same as those used in the simulated environment, simplifying the impact of initial poses on the pre-trained model. As discussed in Section \ref{IIIC}, to mitigate occlusion of the end-effector and rolling pin from the first-person view camera during manipulation, a model-based rollout policy is adopted during training in the simulator. Consequently, the entire manipulation process is divided into multiple rollouts. After each rollout, visual feedback is captured and provided to the policy, which generates the bimanual trajectories for the subsequent rollout based on this feedback. The experimental results are presented in Fig. \ref{real}, showcasing the performance of the deployed policy on the real-world robot.

\section{Discussion}
It is worth noting that this paper only focuses on the two-dimensional shape change of dough. Point cloud information and three-dimensional graph neural network will be introduced, like SoftGPT \cite{liu2023softgpt}, to realize the complete deformation manipulation on a real humanoid robot.

\section{Conclusion}
This paper proposes a dynamic heterogeneous graph-based dough manipulation policy learning model and elaborates its pipeline to deploy the model on a humanoid robot to achieve a dough rolling task with a rolling pin. We show the superiority of using a unified graph form, which can learn the goal-oriented policy and object dynamics and integrate the demonstration guidance simultaneously without the interference of environmental change. Compared with several baseline methods, the proposed DGform with a compact representation and demonstration guidance achieves a human-like dough rolling behavior. The generalizations of these methods are also analyzed, which show the robustness of DGform against environmental change. Finally, a real-world experiment was conducted on a CURI robot, aided by bimanual LQT with coordination.

\bibliographystyle{ieeetr}
\bibliography{ref_dg}

\begin{thebibliography}{10}

\bibitem{yin2021modeling}
H.~Yin, A.~Varava, and D.~Kragic, ``Modeling, learning, perception, and control
  methods for deformable object manipulation,'' {\em Science Robotics}, vol.~6,
  no.~54, p.~eabd8803, 2021.

\bibitem{sun2022mixline}
Z.~Sun, Z.~Wang, J.~Liu, M.~Li, and F.~Chen, ``Mixline: A hybrid reinforcement
  learning framework for long-horizon bimanual coffee stirring task,'' in {\em
  International Conference on Intelligent Robotics and Applications},
  pp.~627--636, Springer, 2022.

\bibitem{liu2022robot}
J.~Liu, Y.~Chen, Z.~Dong, S.~Wang, S.~Calinon, M.~Li, and F.~Chen, ``Robot
  cooking with stir-fry: Bimanual non-prehensile manipulation of semi-fluid
  objects,'' {\em IEEE Robotics and Automation Letters}, vol.~7, no.~2,
  pp.~5159--5166, 2022.

\bibitem{hu2019difftaichi}
Y.~Hu, L.~Anderson, T.~Li, Q.~Sun, N.~Carr, J.~Ragan{-}Kelley, and F.~Durand,
  ``Difftaichi: Differentiable programming for physical simulation,'' in {\em
  8th International Conference on Learning Representations, {ICLR} 2020, Addis
  Ababa, Ethiopia, April 26-30, 2020}, OpenReview.net, 2020.

\bibitem{RoboCraft}
H.~Shi, H.~Xu, Z.~Huang, Y.~Li, and J.~Wu, ``Robocraft: Learning to see,
  simulate, and shape elasto-plastic objects with graph networks,'' 2022.

\bibitem{schaal2006dynamic}
S.~Schaal, ``Dynamic movement primitives-a framework for motor control in
  humans and humanoid robotics,'' in {\em Adaptive motion of animals and
  machines}, pp.~261--280, Springer, 2006.

\bibitem{calinon2016tutorial}
S.~Calinon, ``A tutorial on task-parameterized movement learning and
  retrieval,'' {\em Intelligent service robotics}, vol.~9, no.~1, pp.~1--29,
  2016.

\bibitem{ho2016generative}
J.~Ho and S.~Ermon, ``Generative adversarial imitation learning,'' {\em
  Advances in neural information processing systems}, vol.~29, 2016.

\bibitem{savva2019habitat}
M.~Savva, J.~Malik, D.~Parikh, D.~Batra, A.~Kadian, O.~Maksymets, Y.~Zhao,
  E.~Wijmans, B.~Jain, J.~Straub, J.~Liu, and V.~Koltun, ``Habitat: {A}
  platform for embodied {AI} research,'' in {\em 2019 {IEEE/CVF} International
  Conference on Computer Vision, {ICCV} 2019, Seoul, Korea (South), October 27
  - November 2, 2019}, pp.~9338--9346, {IEEE}, 2019.

\bibitem{todorov2012mujoco}
E.~Todorov, T.~Erez, and Y.~Tassa, ``Mujoco: A physics engine for model-based
  control,'' in {\em 2012 IEEE/RSJ international conference on intelligent
  robots and systems}, pp.~5026--5033, IEEE, 2012.

\bibitem{huang2021plasticinelab}
Z.~Huang, Y.~Hu, T.~Du, S.~Zhou, H.~Su, J.~B. Tenenbaum, and C.~Gan,
  ``Plasticinelab: A soft-body manipulation benchmark with differentiable
  physics,'' in {\em International Conference on Learning Representations
  (ICLR)}, 2020.

\bibitem{liu2020efficient}
J.~Liu, J.~Shou, Z.~Fu, H.~Zhou, R.~Xie, J.~Zhang, J.~Fei, and Y.~Zhao,
  ``Efficient reinforcement learning control for continuum robots based on
  inexplicit prior knowledge,'' {\em arXiv preprint arXiv:2002.11573}, 2020.

\bibitem{figueroa2016learning}
N.~Figueroa, A.~L.~P. Ureche, and A.~Billard, ``Learning complex sequential
  tasks from demonstration: A pizza dough rolling case study,'' in {\em 2016
  11th ACM/IEEE International Conference on Human-Robot Interaction (HRI)},
  pp.~611--612, Ieee, 2016.

\bibitem{calinon2013improving}
S.~Calinon, T.~Alizadeh, and D.~G. Caldwell, ``On improving the extrapolation
  capability of task-parameterized movement models,'' in {\em 2013 IEEE/RSJ
  International Conference on Intelligent Robots and Systems}, pp.~610--616,
  IEEE, 2013.

\bibitem{lin2022diffskill}
X.~Lin, Z.~Huang, Y.~Li, J.~B. Tenenbaum, D.~Held, and C.~Gan, ``Diffskill:
  Skill abstraction from differentiable physics for deformable object
  manipulations with tools,'' {\em arXiv preprint arXiv:2203.17275}, 2022.

\bibitem{radford2019language}
A.~Radford, J.~Wu, R.~Child, D.~Luan, D.~Amodei, I.~Sutskever, {\em et~al.},
  ``Language models are unsupervised multitask learners,'' {\em OpenAI blog},
  vol.~1, no.~8, p.~9, 2019.

\bibitem{liu2023softgpt}
J.~Liu, Z.~Li, W.~Lin, S.~Calinon, K.~C. Tan, and F.~Chen, ``Softgpt: Learn
  goal-oriented soft object manipulation skills by generative pre-trained
  heterogeneous graph transformer,'' in {\em 2023 IEEE/RSJ International
  Conference on Intelligent Robots and Systems (IROS)}, pp.~4920--4925, IEEE,
  2023.

\bibitem{weng2021graph}
Z.~Weng, F.~Paus, A.~Varava, H.~Yin, T.~Asfour, and D.~Kragic, ``Graph-based
  task-specific prediction models for interactions between deformable and rigid
  objects,'' in {\em 2021 IEEE/RSJ International Conference on Intelligent
  Robots and Systems (IROS)}, pp.~5741--5748, IEEE, 2021.

\bibitem{kipf2016semi}
T.~N. Kipf and M.~Welling, ``Semi-supervised classification with graph
  convolutional networks,'' {\em arXiv preprint arXiv:1609.02907}, 2016.

\bibitem{kumar2020conservative}
A.~Kumar, A.~Zhou, G.~Tucker, and S.~Levine, ``Conservative q-learning for
  offline reinforcement learning,'' in {\em Advances in Neural Information
  Processing Systems 33: Annual Conference on Neural Information Processing
  Systems 2020, NeurIPS 2020, December 6-12, 2020, virtual} (H.~Larochelle,
  M.~Ranzato, R.~Hadsell, M.~Balcan, and H.~Lin, eds.), 2020.

\bibitem{liu2023birp}
J.~Liu, H.~Sim, C.~Li, K.~C. Tan, and F.~Chen, ``Birp: Learning robot
  generalized bimanual coordination using relative parameterization method on
  human demonstration,'' in {\em 2023 62nd IEEE Conference on Decision and
  Control (CDC)}, pp.~8300--8305, IEEE, 2023.

\bibitem{calinon2017learning}
S.~Calinon and D.~Lee, ``Learning control,'' in {\em Humanoid robotics: A
  reference}, pp.~1--52, Springer Netherlands, 2017.

\bibitem{liu2023rofunc}
J.~Liu, Z.~Dong, C.~Li, Z.~Li, M.~Yu, D.~Delehelle, and F.~Chen, ``Rofunc: The
  full process python package for robot learning from demonstration and robot
  manipulation,'' 2023.

\end{thebibliography}

\end{document}